\documentclass[runningheads]{llncs}
\usepackage{amssymb}
\usepackage{graphicx}
\usepackage{float}
\usepackage{url}
\newcommand{\tabincell}[2]{\begin{tabular}{@{}#1@{}}#2\end{tabular}}  

\begin{document}
\title{CFUN: Combining Faster R-CNN and U-net Network for Efficient Whole Heart Segmentation}
\titlerunning{\ }
%
\author{Zhanwei Xu \and Ziyi Wu \and Jianjiang Feng}
\authorrunning{Zhanwei Xu et al.}
%
\institute{
\email{xzw14@tsinghua.org.cn}\\
\email{wuzy17@mails.tsinghua.edu.cn}\\
\email{jfeng@mail.tsinghua.edu.cn}
}
\maketitle              
\begin{abstract}
In this paper, we propose a novel heart segmentation pipeline Combining Faster R-CNN and U-net Network (CFUN). Due to Faster R-CNN's precise localization ability and U-net's powerful segmentation ability, CFUN needs only one-step detection and segmentation inference to get the whole heart segmentation result, obtaining good results with significantly reduced computational cost. Besides, CFUN adopts a new loss function based on edge information named 3D Edge-loss as an auxiliary loss to accelerate the convergence of training and improve the segmentation results. Extensive experiments on the public dataset show that CFUN exhibits competitive segmentation performance in a sharply reduced inference time. Our source code and the model are publicly available at \url{https://github.com/Wuziyi616/CFUN}.

\end{abstract}
\section{Introduction}
The whole heart segmentation of cardiac CT volumes can be applied in many medical fields~\cite{frangi2001three}~\cite{ecabert2008automatic}~\cite{zheng2008four}~\cite{ecabert2011segmentation} and has received renewed interests since the deep convolutional neural network (DCNN) becomes popular~\cite{payer2017multi}~\cite{yang20173d}~\cite{wang2018two}. However, some crucial challenges remain. Because of the memory limit of GPU, a tiling strategy is usually adopted~\cite{Ronneberger2015U}, namely, the segmentation network needs to process small patches of the volume and stack them together to form the final heart segmentation result. Such strategy fails to capture a global constraint of heart anatomy and wastes lots of computing resources since the heart volume only occupies a small part of the whole CT image. 

We propose a detection-based segmentation method so that the segmentation network can focus on the most relavant part of the CT image and does not need to divide the whole image into patches. Another advantage is that the segmentation network takes the whole heart image enclosed by the RoI as input so it can use the information of the heart as a whole to get a more regularized segmentation result.

While the detection and localization network of CFUN is based on Faster R-CNN~\cite{ren2017faster} and the segmentation network is based on 3D U-net~\cite{cciccek20163d}, we have made some important modifications. Unlike the original Faster R-CNN~\cite{ren2017faster} which proposes many boxes of different objects, we force the detection network to recommend no more than one bounding box containing the whole heart's anatomical structures. The ResNet structure in Faster R-CNN is replaced with P3D Bottleneck~\cite{qiu2017learning} considering the imaging characteristics of CT images and a 3D Edge-loss head is chosen as an auxiliary task for 3D U-net to get more precise segmentation. More detailed modifications of network structures will be discussed in Section 3.
\begin{figure}
\includegraphics[width=\textwidth]{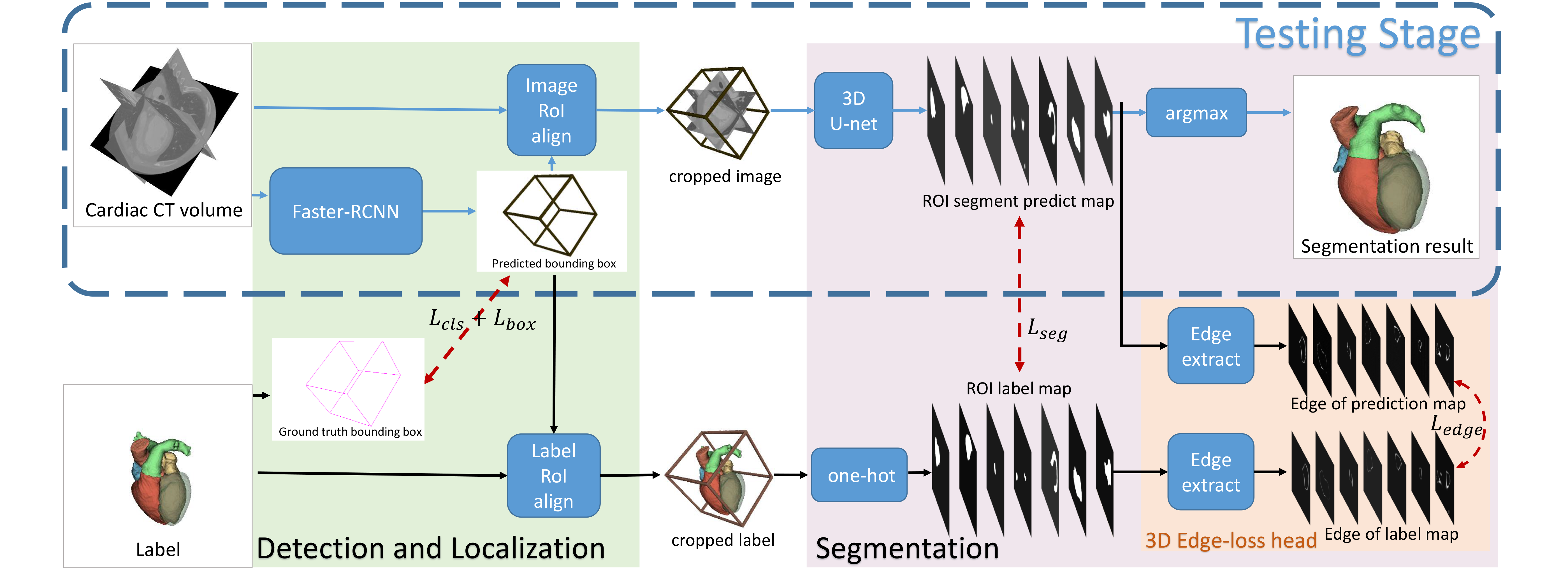}
\caption{Pipeline of CFUN. The heart detection and localization part outputs an RoI containing the whole heart and the following segmentation network segments 8 structures at one step. The 3D Edge-loss head is used in the training stage to help the segmentation network achieve better performance at the boundaries of the anatomical structures. Only the part contained in the dotted box is required during the test.} \label{Figure 1}
\end{figure}

\section{Related work}
\subsection{Traditional Methods}
Model-based and atlas-based image analysis methods have been widely used for extracting parameters of cardiac shape and function from 3-dimensional cardiac images~\cite{frangi2001three}. Ecabert et al.~\cite{ecabert2008automatic}~\cite{ecabert2011segmentation} first locate the heart using Generalized Hough Transformation and then adapt heart chambers' model to fit images. The fitting degree of the model can be optimized by minimizing a designed energy function. Unlike~\cite{ecabert2008automatic} and~\cite{ecabert2011segmentation}, Zheng et al.~\cite{zheng2008four} exploit important landmarks among the control points of the model to guide the automatic model fitting process. After localizing the heart chambers, learning based boundary delineation is used to form the segmentation shape. Zhuang et al.~\cite{zhuang2010registration}  propose a atlas-based heart segmentation method consisting of two image registration algorithms, one is to match anatomical substructures and the other one is to refine local details.

These traditional methods usually have one or more of the following limitations: 1) difficult to generalize to segmentation of other organs, 2) difficult to utilize the computational power of modern GPU, 3) manually designed features are usually not optimal.

\subsection{3D U-net Based Segmentation Network}
Fully Convolutional Network (FCN)~\cite{long2015fully} provides a new way to solve the semantic segmentation task of natural images, which uses an end-to-end full convolution structure and can segment images of arbitrary size. By adding skipping short-link structure, U-net~\cite{Ronneberger2015U} network achieves better segmentation performance than FCN, and gradually becomes the mainstream in the field of medical image segmentation. Because of the increase of storage and computing power of GPU, 3D U-net directly using 3D volume as input is also gaining popularity~\cite{cciccek20163d}. The 3D U-net structure is also the backbone of out segmentation subnet.

Despite its excellent performance, 3D U-net often needs to be used together with tiling strategy. Specifically, we cut the 3D volume into small patches in order, enter them into the network separately to get the output, and then stitch the output together to get the final segmentation result. Considering that most parts of the heart images belong to the background, this strategy obviously wastes computational resources. 

Researchers have made some efforts to deal with the problem. Wang et al.~\cite{wang2018two} use a two-stage 3D U-net framework where the first U-net segments a coarse resolution heart image and the second one takes the original image plus the output of the first U-net as input and outputs segmentation of original resolution. This strategy makes the quality of final segmentation heavily rely on the quality of the coarse segmentation result, and the error of the coarse segmentation result will spread to the original segmentation. Also, the second U-net uses 2D conv kernels in the decoding path and thus its ability to integrate axial information is not as strong as the strategy of integrating the other two dimensions’ information. A two-stage heart segmentation approach is proposed by Payer et al.~\cite{payer2017multi}, who also adopt a coarse-to-fine method. Instead of directly regressing the size and coordinates of the bounding box, they regress the center of the heart and then assume a bounding box of fixed size, which affects the accuracy of segmentation.

\subsection{Faster R-CNN and Region-proposal Based Segmentation}
After two improvements~\cite{girshick2015fast}~\cite{girshick2014rich}, Faster R-CNN~\cite{ren2017faster} shows excellent detection and positioning capability and is very efficient. Pictures are passed through an RPN (region proposal network) first to get many recommendation boxes and then a classification and regression network is directly added to the resized corresponding features of the recommendation boxes to get the classification result and the final bounding box of the object.

By adding a mask head after the RPN, Mask R-CNN~\cite{he2017mask} network can simultaneously perform detection, classification, and segmentation. There are already some attempts to apply Mask R-CNN to medical image segmentation. A direct migration comes from Zhao et al.~\cite{zhao2018deep}, who apply it to instance segmentation of developing embryos. Compared with the segmentation of developing embryos, whole heart segmentation task is much more complicated and the segment head of Mask R-CNN cannot handle. As shown in our ablation experiment, the performance of Mask R-CNN on heart segmentation is not as good as that of our proposed method.


\section{Method}
CFUN consists of two parts in sequence: a modified 3D Faster R-CNN to propose a bounding box centering the heart, a 3D U-net to segment the heart-centered image with an Edge-loss head to calculate the edge information and compare it with the real edge. During the test, the Edge-loss head is not needed. Fig.~\ref{Figure 1} shows our pipeline and we will discuss more details of the method in the following sections.
\subsection{Heart Detection and Localization}
We extend the Faster R-CNN network to 3 dimensions by replacing all the 2D kernels with 3D kernels. Just like the video data where the frame dimension is different from the images' dimension~\cite{qiu2017learning}, the resolution of $z$-axis of cardiac CT volume is different from the other two axes. As a result, the ResNet~\cite{he2016deep} part of the Faster R-CNN in the original paper~\cite{ren2017faster} is replaced by a P3D ResNet structure~\cite{qiu2017learning} which has been proved beneficial in video analysis. The P3D network divides the 3 dimensions into two orthogonal bases and simulates $3 \times 3 \times 3$ convolutions with $1 \times 3 \times 3$ convolutional filters plus $3 \times 1 \times 1$ convolutions. We also add FPN (Feature Pyramid Networks)~\cite{lin2017feature} after the P3D ResNet to combine the feature maps in different resolutions. These two subnets comprise the RPN part of our method.

As we embed the 3D U-net into the modified Faster R-CNN, the original bounding box refine head still remains. Our ablation experiments will show that the remaining head contributes to the following segmentation process.

Compared with the method that directly regresses the bounding box of a heart, our Region Proposal Network (RPN) based detection and localization method has the following advantages. First, anchors applied in the RPN network enhance its ability to process multi-scale CT images. Second, bounding box refine head can be easily added to our RPN structure, largely improving the accuracy of ultimate detection and localization results which has been demonstrated by our ablation experiments. Last but not least, our RPN network can be adapted to multiple organ detection and segmentation tasks with little adjustment.

\begin{figure}
\includegraphics[width=\textwidth]{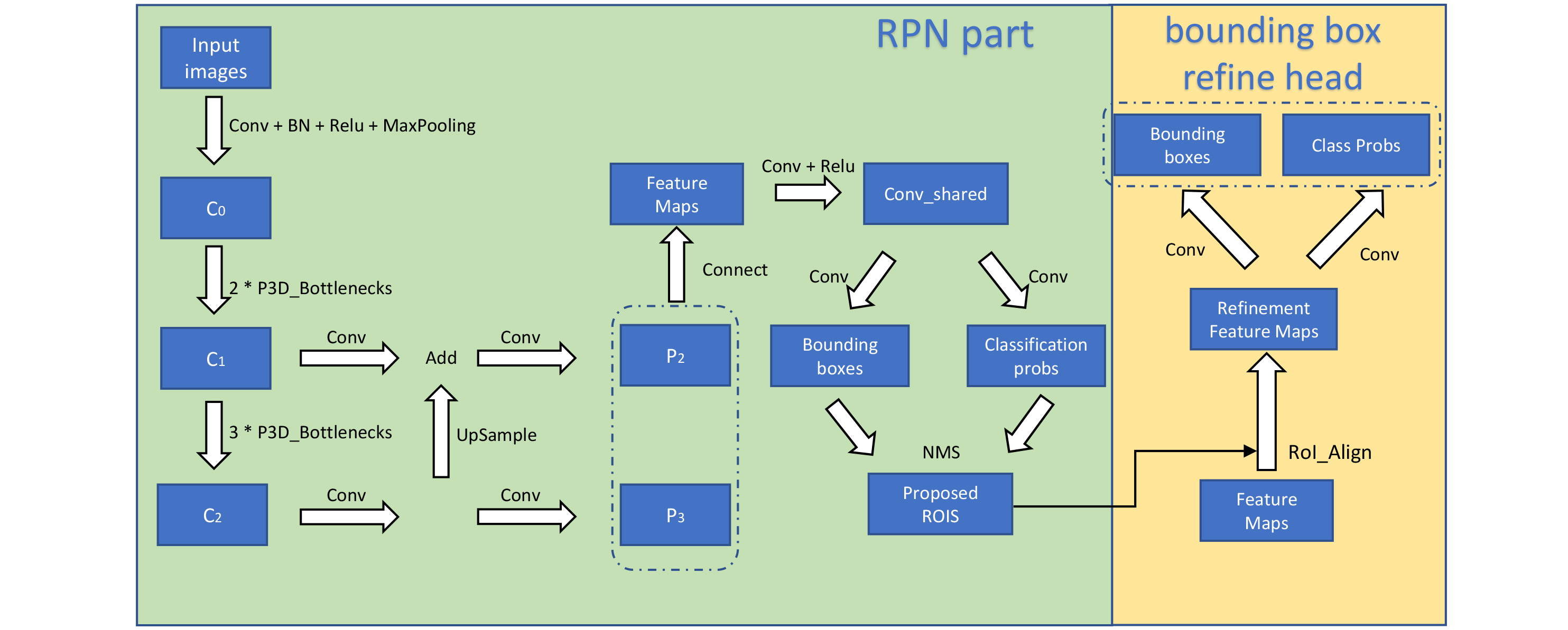}
\caption{Detailed structure of our heart detection and localization network. The main innovation of our RPN part is to adopt P3D bottleneck to better handle different dimensions' information of CT images. Bounding box refine head is really beneficial to the accuracy of detection and localization. The blue boxes in the graph represent features or output; arrows represent operations on features.} \label{Figure 5}
\end{figure}

\subsection{Heart Segmentation}
We modify 3D U-net structure as the segmentation network of CFUN. Instead of aligning the feature maps as input to the segmentation network like Mask R-CNN, we align the original CT images lying in the bounding box, which can be seen as a separation of segmentation head and bounding box refine head of Mask R-CNN. Although He et al.~\cite{he2017mask} prove that we can learn a feature contributing the detection and segmentation at the same time, our ablation experiments still expose its difficulty and inferior performance considering the large RoI-align size ($64\times64\times64$ of ours). Skillfully, we utilize the idea of deep supervision~\cite{lee2015deeply} by adding the output of different resolutions in the decode path as the final prediction output, a variant of \cite{mehta2017m}, and upscale the output size from $64\times64\times64$ to $128\times128\times128$ by adding an extra deconv layer, making up for the loss of precision caused by the downsampling operation.
\begin{figure}
\includegraphics[width=\textwidth]{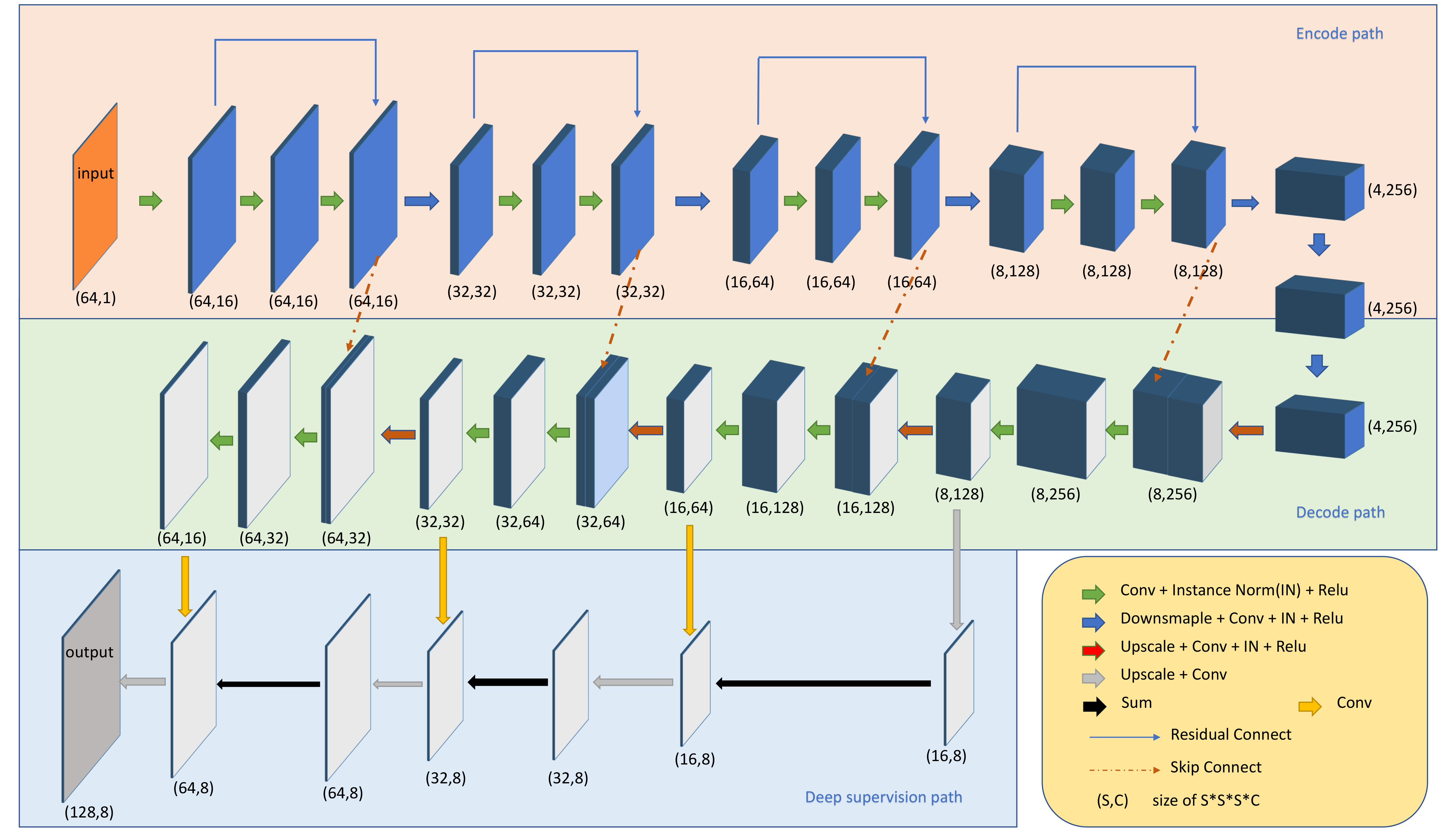}
\caption{Our modified 3D U-net differs from the original 3D U-net in three aspects: 1) residual connections are added in the encode path to learn better feature maps, 2) we add an extra deconv layer to upscale the output so that the output size is twice bigger than input size (RoI-align size of RPN part), 3) we add deep supervision path combining informantion from different resolution to get better segmentation performance. } \label{Figure 3}
\end{figure}

\subsection{Edge-loss Head and Loss Function}
Inspired by the work of Zimmermann et al.~\cite{zimmermann2018faster}, we compensate the accuracy loss by adding an Edge-loss head since the embedded 3D U-net is simplified. ~\cite{zimmermann2018faster} reveals that the Edge-loss head can not only help get better segmentation edge results but also speed up the convergence. Specifically, we apply fixed 3D Sobel-kernels to the final prediction map as well as the one-hot ground truth and calculate the $l_2$ loss between them, called edge loss. The edge loss encourages the 3D U-net to focus more on the edge among heart organs and the background, which is usually more difficult to learn. We design 6 3D Sobel-kernels corresponding to 6 directions (positive $x$, $y$, $z$, and negative $x$, $y$, $z$), each kernel can be described as a $3\times3\times3$ matrix where the middle $3\times3$ matrix $S_{middle}$ is zero, the upper $3\times3$ matrix $S_{up}$ is given the values as shown Fig. 4 and the $3\times3$ matrix below $S_{down}$ is given the negative values of the corresponding position in upper $3\times3$ matrix. Edge extraction can be realized by a convolution operation between Sobel kernels and the segmentation predicted map/ground truth. The Edge loss is defined as follows:
\begin{equation}
L_{edge}=\frac{1}{C}\frac{1}{K}\frac{1}{M}\sum_{c=i}^C\sum_{k=1}^K\sum_{m=1}^M ||E_k(y)^c_m-E_k(p)^c_m||_2
\end{equation}
where $C$ is the classes of segmentation, $K$ is the Sobel kernels' number, $M$ is the voxel number of the image, $y$ is the one-hot label map and $p$ is the prediction map of the network. $E_k(y)^c_m$ means the $m$-th pixel value after extracting the $k$-th edge information from the $c$-th dimension of the image $y$. 
\begin{figure}
\includegraphics[width=\textwidth]{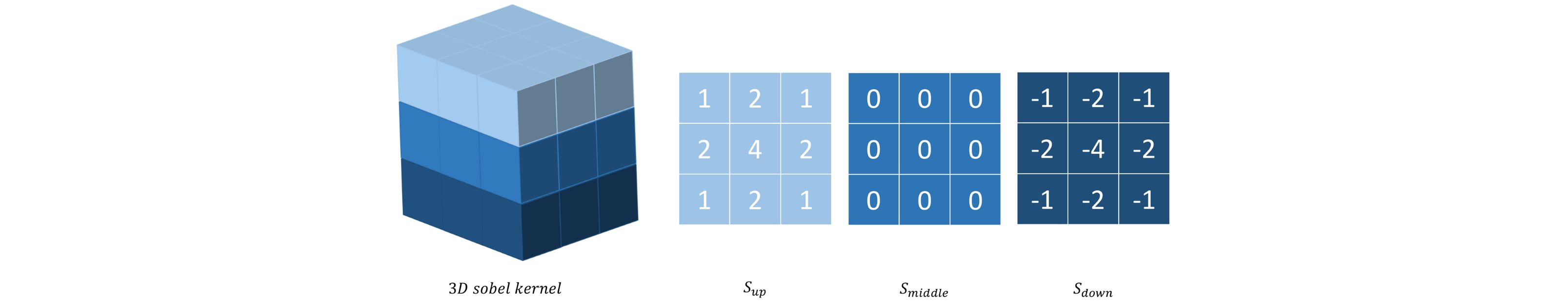}
\caption{One of 3D Sobel kernels' visualization.} \label{Figure 2}
\end{figure}

CFUN optimizes four loss functions: segmentation loss of positive bounding box $L_{seg}$, edge loss of edge head $L_{edge}$, regression loss and classification loss of bounding boxes $L_{box}$ and $L_{cls}$. The definition and calculation of $L_{seg}$, $L_{cls}$ and $L_{box}$ are the same as those in Faster R-CNN~\cite{ren2017faster} and Mask R-CNN~\cite{he2017mask}, while $L_{edge}$ is the $L_2$ distance between the edges of the predicted map and the labeled segmentation as defined as (1), so the whole Loss can be identified as:
\begin{equation}
Loss = w_1 L_{box}+w_2 L_{cls}+w_3 L_{seg}+w_4 L_{edge}
\end{equation}
where $w_1$, $w_2$, $w_3$ and $w_4$ are parameters used to balance the contribution of different losses. In our experiments, we set $w_1:w_2:w_3:w_4=2:2:2:1$.

\section{Experiments}
\subsection{Dataset}
All the 60 heart CT images are from the MM-WHS2017 challenge~\cite{zhuang2016multi}~\cite{zhuang2013challenges}~\cite{zhuang2010registration}, containing 20 training volumes and 40 testing volumes. The labels contain 7 anatomical structures including the left ventricle blood cavity (LV), the myocardium of the left ventricle (Myo), the right ventricle blood cavity (RV), the left atrium blood cavity (LA), the right atrium blood cavity (RA), the ascending aorta (AA) and the pulmonary artery (PA). Since only the 20 training samples' segmentation annotations are public and our network requires more training samples to avoid overfitting, we have to redivide the data set. We use 40 original test samples as training samples, 5 random selected original training samples as validation sets, and the remaining 15 original training samples as test sets. An automated algorithm is designed to generate segmentation labels of 40 original test samples although the segmentation results are not as accurate as manual annotations, which affects the test performance of CFUN. The above database redivided method also ensures that the results of different algorithms can be compared fairly in the open manual segmentation dataset. 

In order to generate the segmentation labels of the testing data, we design a 3D U-net network with deep supervision plus weighted loss function, train it using detached training data patches, and segment the testing data by tiling strategy. We submit our testing segmentation results and the organizers of MM-WHS2017 challenge feedback the average Dice score is 0.88.

\subsection{Training strategy and detail}
The proposed network is implemented in PyTorch, using an NVIDIA GeForce GTX 1080Ti GPU. We resize the CT images to $320\times320\times192$ as input and the RoI-align size is $64\times64\times64$. We update the weights of the network with an Adam optimizer (batch size=1, learning rate is 0.001, steps of each epoch is 32). Although the Edge-loss head can reduce the iteration in the training stage, the time cost of one iteration becomes longer with it. We train the network without the Edge-loss head in the first 300 epochs, and add the Edge-loss head in the next 100 epochs as a fine-tuning, so the total iteration is $\sim$12000.

As for the hyper-parameter of the Faster R-CNN network, detection threshold of the bounding box is 0.5, negative and positive bounding box ratio is 1:2, anchor ratio is fixed as 1:1:1, anchor scales are 64 and 128 to fit the different heart ratio in CT images.

\subsection{Testing results}
\begin{figure}
\includegraphics[width=\textwidth]{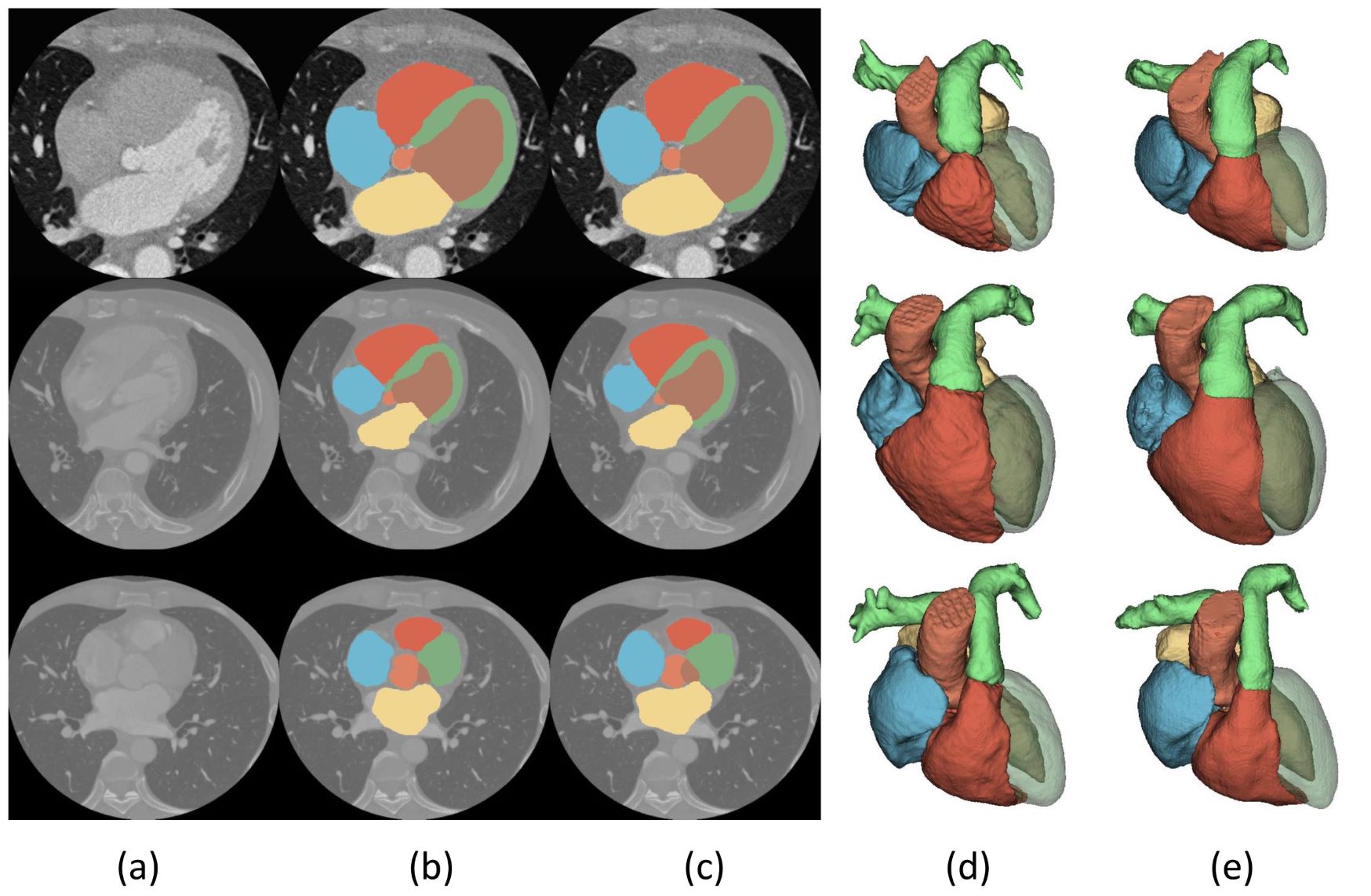}
\caption{Visualization of some test results. From top to bottom, the three CT images are 1007, 1009, and 1019 respectively. (a) shows the original CT images, (b) shows ground truth, (c) shows the predicted segmentation, (d) and (e) show 3D visualization of ground truth and predicted segmentation respectively. Better viewed in color.} \label{Figure 4}
\end{figure}
Without bells and whistles, CFUN reaps an average 85.9\% Dice score on test as shown in Table~\ref{Table1} even though our training ground truth isn't that precise. The performance is better than 3D U-net method in MM-WHS2017~\cite{yang20173d} and two-stage 3D U-net method in MICCAI2018~\cite{wang2018two}, comparable to the winner of the MM-WHS2017~\cite{payer2017multi}. What's more exciting is the speed. It only takes less than 10 seconds for our network to segment a heart and less than 15 seconds if all data-loading and post-processing operations are taken account. Compared with tiling strategy or other methods (usually spend more than several minutes), it is a great improvement.
\begin{table}
\centering
\caption{Test results on MM-WHS2017. CFUN takes less than 15 seconds to segment a heart. Note that our training ground truth isn't as precise as manually labelled ones.}\label{Table1}
\begin{tabular}{l|cccccccc}
\hline
&LV&	Myo	&RV	&LA	&RA	&AA	&PA	&Average\\
\hline
 3D U-net~\cite{yang20173d}&0.813&0.791&0.909&0.853&0.816&0.717&0.763&0.809\\
 two stage U-net~\cite{wang2018two}&0.799&0.729&0.786&0.904&0.793&0.873&0.648&0.806\\
 SEG-CNN~\cite{payer2017multi}&0.924&0.872&0.879&0.910&0.865&0.940&0.837&0.887\\
 CFUN&0.879&0.822&0.902&0.832&0.844&0.913&0.821&0.859\\
\hline
\end{tabular}
\end{table}

\section{Ablation experiments}
In order to better study and illustrate the contribution of different structures of CFUN to segmentation performance, we conducted a series of ablations. Note that all the comparative experiments follow the strategy and evaluation criteria mentioned in the experiment section.
\subsection{Different feature maps as the input of 3D U-net}
We can use the aligned original image I(O), the aligned corresponding conv layers I(C), or the aligned corresponding feature maps I(P) lying in the RoI of the Faster R-CNN as input of the 3D U-net. If we chose I(P), the structure will be more like Mask R-CNN. We tried these three align methods respectively and the test results are shown in Table~\ref{Table2}. Feature maps in Faster R-CNN with FPN structure have skipping connections and encode-decode path similar to U-net, making I(P)'s performance better than I(O)'s. Beyond our expectations, I(P)'s performance didn't exceed I(O)'s. Some factors account for this. The first reason is our RoI-align size ($64\times64\times64$) is much bigger than that in the original Mask R-CNN~\cite{he2017mask} ($14\times14$). Another reason is that our segmentation network must segment several anatomical structures from one bounding box, while in the original network~\cite{he2017mask}, only one foreground object is contained in each bounding box. These two things increase the difficulty to learn a feature map that contributes both the segmentation task and the detection task. 

\begin{table}
\centering
\caption{Test results with different inputs to the segmentation subnet. Note that we didn't add Edge-loss head in this experiment.}\label{Table2}
\begin{tabular}{l|cccccccc}
\hline
&LV&	Myo	&RV	&LA	&RA	&AA	&PA	&Average\\
\hline
I(C)&0.707&0.539&0.701&0.665&0.653&0.704&0.522&0.641\\
I(P)&0.732&0.573&0.732&0.635&0.673&0.718&0.564&0.661\\
I(O)&0.809&0.639&0.801&0.739&0.747&0.785&0.573&0.727\\
\hline
\end{tabular}
\end{table}

\subsection{w/o Edge-loss head and w/o bounding box refine head}
\begin{figure}
\includegraphics[width=\textwidth]{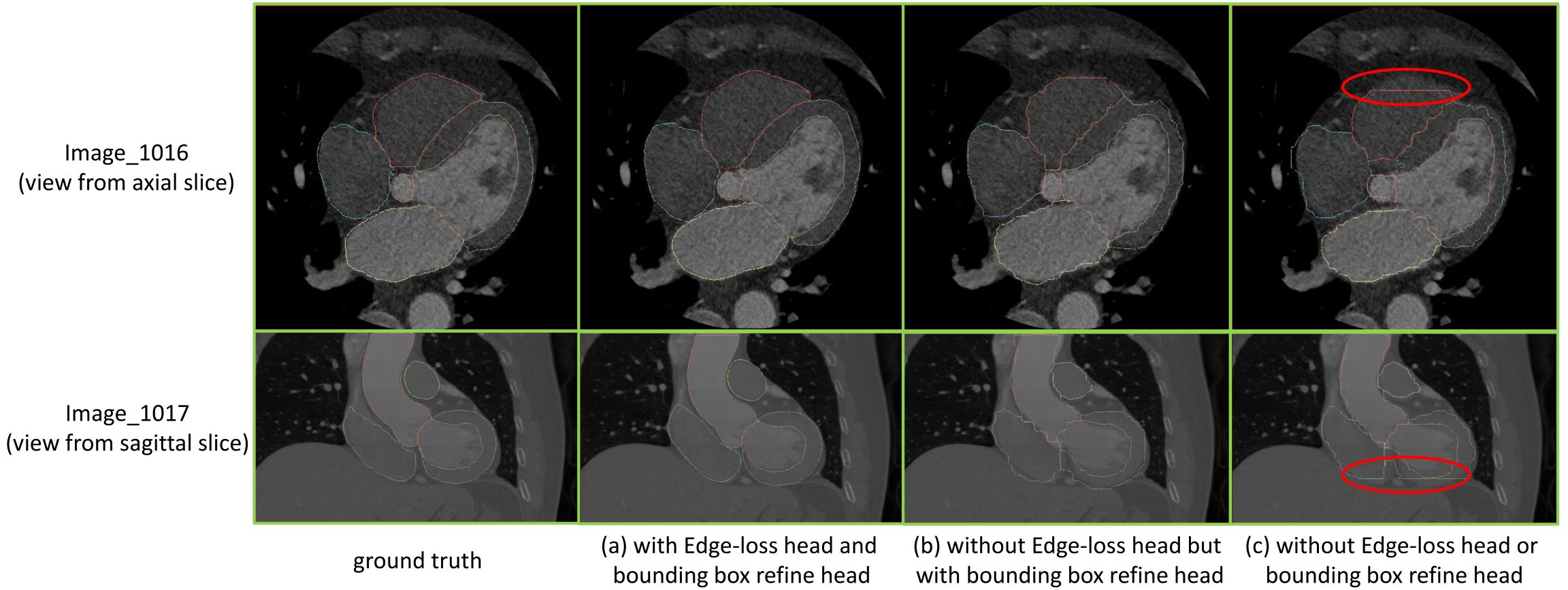}
\caption{Visualization of segmentation results. Lines represent contours of different anatomical structures in segmentation results. Comparing (a) and (b), we can see the Edge-loss head helps the performance at the edge; Comparing (b) and (c), we can see the bounding box refine head improves the detection precision as the red circle marks. Better viewed in color.} \label{Figure 6}
\end{figure}

For a fair comparison, we force the total iteration w/o Edge-loss head and w/o bounding box refine head to be the same (12000). The test result is shown in Table~\ref{Table3}. and a visualization of segmentation result is shown in Fig~\ref{Figure 6}. Notice that the bounding box refine head improves the detection precision and the Edge-loss head benefits the quality of the anatomical structure's edges.
\begin{table}
\centering
\caption{Test results w/o Edge-loss head and w/o bounding box refine head.}\label{Table3}
\begin{tabular}{cc|cccccccc}
\hline
edge loss&\ \ \tabincell{c}{bounding box\\refine head}& LV&	Myo	&RV	&LA	&RA	&AA	&PA	&Average\\
\hline
$\times$ &$\times$&0.793&0.575&0.761&0.648&0.608&0.758&0.462&0.658\\
$\times$&$\checkmark$&0.809&0.639&0.801&0.739&0.747&0.785&0.573&0.728\\
$\checkmark$ &$\checkmark$ &0.879&0.822&0.902&0.832&0.844&0.913&0.821&0.859\\	
\hline
\end{tabular}
\end{table}

\section{Conclusion}
In this paper, we propose a new method Combining Faster R-CNN and U-net Network (CFUN) for the whole heart segmentation task. Our main contribution includes:
\begin{enumerate}
 \item We embed 3D U-net into a modified Faster R-CNN network and thus can quickly segment the whole heart in inference. As far as we know, it's the first time to combine these two networks' structure for whole heart segmentation.
 \item We modify 3D U-net and Faster R-CNN and propose a new connection way to better handle the heart segmentation task.
 \item We design a 3D Edge-loss head to improve the performance of segmentation.
\end{enumerate}
With a very short inference time, CFUN gets average Dice score of 0.859 on MM-WHS2017 dataset, comparable to the winner of the competition. One limitation is that, limited by the GPU's memory, all CT images are downsampled before entered into the network. In the future, we will investigate techniques to maximize the gain of segmentation accuracy using images of original resolution.

\bibliographystyle{splncs04}
\bibliography{mybibliography}

\end{document}